\documentclass{article}

\usepackage[final,nonatbib]{nips_2018}

\usepackage[utf8]{inputenc} 
\usepackage[T1]{fontenc}    
\usepackage{hyperref}       
\usepackage{url}            
\usepackage{booktabs}       
\usepackage{amsfonts}       
\usepackage{nicefrac}       
\usepackage{microtype}      
\usepackage{enumitem}

\usepackage{amsmath}
\usepackage{amssymb}
\usepackage{graphicx}
\usepackage{xcolor}
\usepackage{color}

\title{Turbo Learning for CaptionBot and DrawingBot}

\author{
  Qiuyuan Huang \\
  Microsoft Research\\
   Redmond, WA, USA\\
  \texttt{qihua@microsoft.com} \\
   \And
  Pengchuan Zhang \\
  Microsoft Research\\
  Redmond, WA, USA\\
  \texttt{penzhan@microsoft.com} \\
  \AND
  Dapeng Wu \\
  University of Florida\\
  Gainesville, FL, USA\\
  \texttt{dpwu@ieee.org} \\
  \And
  Lei Zhang \\
  Microsoft Research\\
   Redmond, WA, USA\\
  \texttt{leizhang@microsoft.com} \\
}

\begin{document}

\maketitle

\begin{abstract}
We study in this paper the problems of both image captioning and
text-to-image generation, and present a novel turbo learning
approach to jointly training an image-to-text generator (a.k.a.
CaptionBot) and a text-to-image generator (a.k.a. DrawingBot). The
key idea behind the joint training is that image-to-text
generation and text-to-image generation as dual problems can form
a closed loop to provide informative feedback to each other. Based
on such feedback, we introduce a new loss metric by comparing the
original input with the output produced by the closed loop. In
addition to the old loss metrics used in CaptionBot and
DrawingBot, this extra loss metric makes the jointly trained
CaptionBot and DrawingBot better than the separately trained
CaptionBot and DrawingBot. Furthermore, the turbo-learning
approach enables semi-supervised learning since the closed loop
	can provide pseudo-labels for unlabeled samples. Experimental
results on the COCO dataset demonstrate that the proposed turbo
learning can significantly improve the performance of both
CaptionBot and DrawingBot by a large margin.


\end{abstract}

\section{Introduction}
\label{sec:Introduction}

Due to the breakthrough of deep learning, recent years have
witnessed great progresses in both computer vision and natural
language processing. As a result,  two fundamental problems --
image captioning and text-to-image generation -- that requires
cross modality understanding have also been intensely studied in
the past few years. Image captioning (a.k.a. CaptionBot) is to
generate a meaningful caption for any given input image, whereas
text-to-image (a.k.a. DrawingBot) is to generate a realistic image
for any input sentence. Regardless of their different
implementations, it is interesting to see that the two problems
can be regarded as dual problems as both can take each other's
output as input. However, despite their duality in terms of their
input and output forms, the two problems were largely  studied
separately in the past, leaving a significant room for
improvement.

To leverage their duality, for the first time, this paper proposes
a turbo-learning approach, which can jointly train a CaptionBot
and a DrawingBot together in a way similar to an engine
turbo-charger, which feeds the output back to the input to reuse
the exhaust gas for increased back pressure and better engine
efficiency. The key idea of the proposed turbo-learning approach
is that an image-to-text generator and a text-to-image generator
can form a closed loop and generate informative feedback signals
to each other. In this way, we can obtain a new loss metric (for
updating the neural network weights during training) by comparing
the original input data with the output data produced by the whole
closed loop, in addition to the old loss metric which measures the
difference between the output and the ground-truth of the
CaptionBot or the DrawingBot. This extra loss metric effectively
leads to improved performance of both of the jointly trained
CaptionBot and DrawingBot compared with their separately trained
counterparts.

To jointly train a CaptionBot and a DrawingBot, we utilize the
state-of-the-art long short-term memory (LSTM) image captioner
\cite{SCN_CVPR2017} and text-to-image generation algorithm
\cite{Tao18attngan} as building blocks and use stochastic gradient
descent to iteratively learn the  network parameters of both
blocks. More specifically, as illustrated in Fig. \ref{fig:turbo},
training in each iteration consists of two steps. In Step 1, the
DrawingBot serves as the primal module while the CaptionBot serves
as the dual module. In Step 2, the CaptionBot serves as the primal
module while the DrawingBot serves as the dual module. In each
step, the primal module takes an input and produces/forwards its
output to its dual module, and the dual module further feeds back
its output to the input of the primal module. The weights of the
DrawingBot and the CaptionBot are updated simultaneously in each
step by using the loss function of two pairs -- (gold image vs.
generated image) and (gold sentence vs. generated sentence) to
calculate the gradient.

Closing the loop between CaptionBot and DrawingBot also brings
another merit: it enables semi-supervised learning since the
closed loop can provide pseudo-labels for unlabeled samples. For
image captioning, such a semi-supervised learning capability is
particularly useful since human annotation of images is very
costly. Note that between the two bots in each step we also have a
constraint to ensure the sentence (or image) generated from the
primal module is natural (or realistic). The constraint could be a
Generative Adversarial Network (GAN) loss if we only have
unsupervised training data, i.e., images and sentences without
correspondence. In this work, we use the semi-supervised setting
for simplicity and use the ground truth sentence (or image) to
supervise/constrain the sentence (or image) generated from the
primal module, while letting the unlabeled data to pass through
the two modules to improve the feature learning. This loss term
effectively prevents the model from learning trivial identity
mappings for both the CaptionBot and DrawingBot.

We conducted experimental evaluation on the COCO dataset.
Experimental results show that the turbo learning approach
significantly improves the performance of CaptionBot and
DrawingBot.  For example, under supervised learning, the
CaptionBot is improved by 13\% in BLEU-4 metric, and the
DrawingBot is improved by 4.3\% in Inception score; under
semi-supervised learning, the CaptionBot is improved by 77\% in
BLEU-4 metric.


The rest of the paper is organized as follows.
Section~\ref{sec:RelatedWork} discusses the related work. In
Section~\ref{sec:TurboLearning}, we introduce our proposed turbo
learning approach and present its theoretical underpinning.
Section~\ref{sec:supervised training} describes turbo learning for
supervised training and presents experimental results.  In
Section~\ref{sec:Semi-supervised}, we present our semi-supervised
learning for CaptionBot and experimental results. Finally,
Section~\ref{sec:Conclusion} concludes the paper.

\section{Related work}
\label{sec:RelatedWork}

Most existing image captioning systems exploit end-to-end deep
learning with a convolutional neural network (CNN) image-analysis front end producing a
distributed representation that is then used to drive a
natural-language generation process, typically using a recurrent neural network (RNN) or LSTM
\cite{mao2015deep,vinyals2015show,devlin2015language,chen2015mind,donahue2015long,karpathy2015deep,kiros2014multimodal,kiros2014unifying,xu2017image,Rennie2016Self,yao2016boosting,lu2017knowing}.



Generating photo-realistic images from text descriptions has been an active research area in recent years. There are different approaches working toward this grand goal, such as variational inference~\cite{MansimovPBS15, Gregor15DRAW}, approximate Langevin process~\cite{Reed17parallel}, conditional PixelCNN via maximal likelihood estimation~\cite{Oord16, Reed17parallel}, and conditional generative adversarial networks~\cite{reed2016generative, reed2016learning, Han16stackgan, Han17stackgan2}. Compared with other approaches, generative adversarial networks (GANs)~\cite{goodfellow2014generative} have shown great performance for generating sharper samples~\cite{Radford15, DentonCSF15, Salimans2016, Christian2016, pix2pix2017}. AttnGAN~\cite{Tao18attngan} proposes an attentional multi-stage generator that can synthesize fine-grained details at different image regions by paying attentions to the relevant words in the text description, which achieves the state-of-the-art performance of the text-to-image synthesis task on MS-COCO dataset.

Different from previous works which study the problems of image
captioning and text-to-image generation separately, this work aims
at developing a turbo learning approach to jointly training
CaptionBot and DrawingBot. Our proposed turbo learning is similar
to dual learning \cite{Xia2016DualLearning}, which is applied to
automatic machine translation. Under dual learning, any machine
translation task has a dual task, e.g., English-to-French
translation (primal) versus French-to-English translation (dual);
the primal and dual tasks can form a closed loop, and generate
feedback signals to train the translation models, even if without
the involvement of a human labeler. In the dual-learning
mechanism,  one agent represents the model for the primal task and
the other agent represents the model for the dual task, then the
two agents teach each other through a reinforcement learning
process. Dual-learning is solely developed for NLP.  In contrast,
our turbo learning is applied to both NLP and computer vision,
which is more challenging.

\section{Turbo learning structure}
\label{sec:TurboLearning}

In this section, we present the turbo learning approach for
training CaptionBot and DrawingBot jointly.

\subsection{Turbo learning for CaptionBot and DrawingBot}
\begin{figure}[ht]
    \centering
    \includegraphics[width=0.6\textwidth]{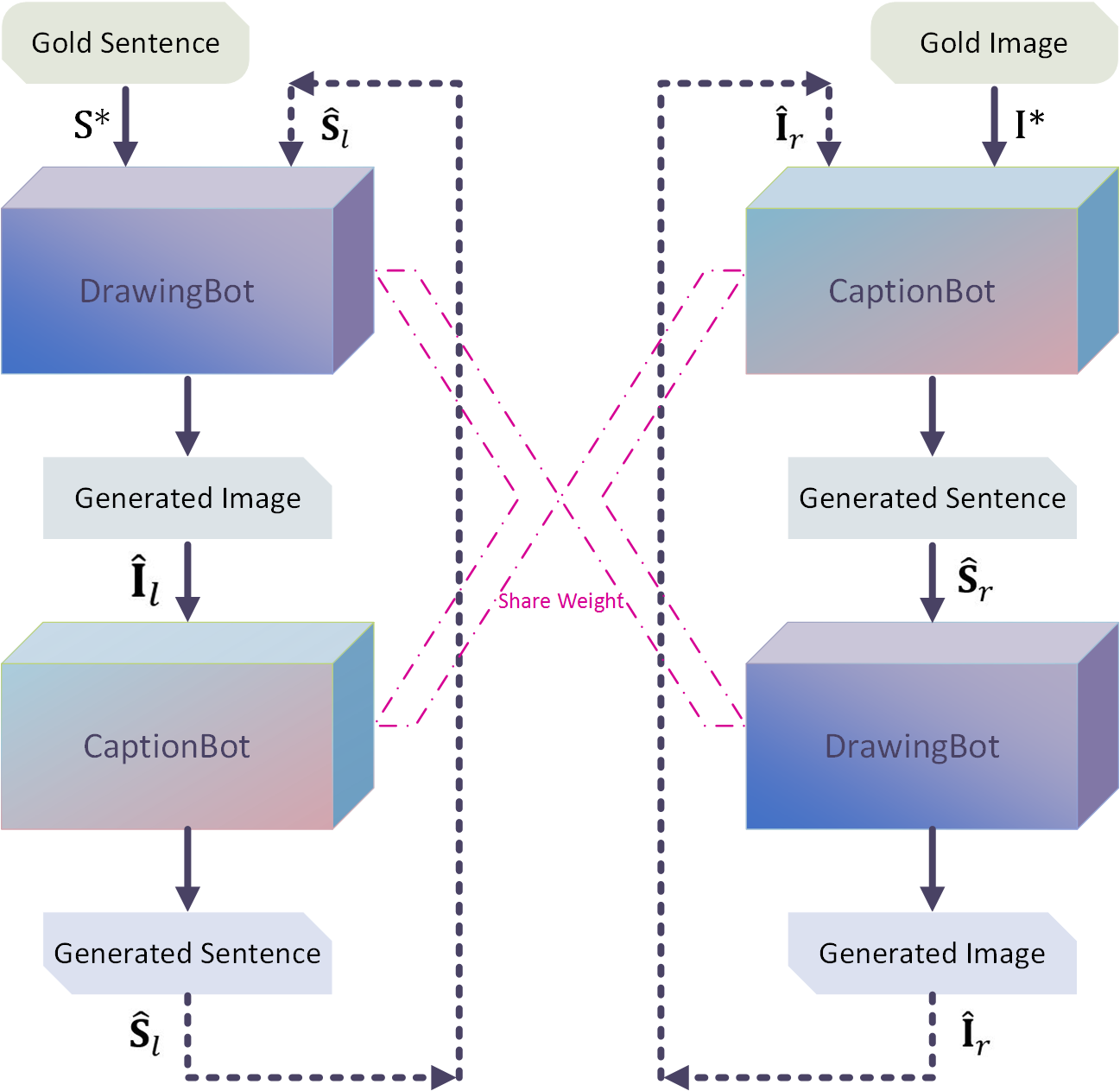}
    \caption{The proposed turbo butterfly structure for training CaptionBot and DrawingBot jointly}
    \label{fig:turbo}
\end{figure}

Fig.~\ref{fig:turbo} shows the turbo architecture for the proposed
joint training of CaptionBot and DrawingBot. As shown in the left
hand side of Fig.~\ref{fig:turbo}, for a given gold sentence
${\mathbf S}^*$, the DrawingBot generates an image $\hat{{\mathbf
I}}_l$; the generated image is supplied to the CaptionBot, which
produces a captioning sentence $\hat{{\mathbf S}}_l$. Then we use
the loss function of pairs $(\hat{{\mathbf S}}_l,{\mathbf S}^*)$
and $(\hat{{\mathbf I}}_l,{\mathbf I}^*)$ to calculate the
gradients and update $\theta_{draw}$ and $\theta_{cap}$, the
parameters of the DrawingBot and CaptionBot, simultaneously.  As
shown in the right hand side of Fig.~\ref{fig:turbo}, for a given
gold image ${\mathbf I}^*$, the CaptionBot generates a sentence
$\hat{{\mathbf S}}_r$; the generated sentence is supplied to the
DrawingBot, which produces an image $\hat{{\mathbf I}}_r$. Then we
use the loss function of pairs $(\hat{{\mathbf S}}_r,{\mathbf
S}^*)$ and $(\hat{{\mathbf I}}_r,{\mathbf I}^*)$ to calculate the
gradients and update $\theta_{draw}$ and $\theta_{cap}$
simultaneously.

\subsection{Optimization theoretical underpinning of turbo learning}

Considering the left hand side of Fig.~\ref{fig:turbo}, for a
given gold sentence ${\mathbf S}^*$, the DrawingBot generates an
image $\hat{{\mathbf I}}_l$; mathematically, we can represent this
process by
\begin{eqnarray}
\hat{{\mathbf I}}_l={\cal{D}}({\mathbf S}^*|\theta_{draw})
\end{eqnarray}
where ${\cal{D}}(\cdot)$ denotes the DrawingBot as a nonlinear
operator parameterized by $\theta_{draw}$. The generated image
$\hat{{\mathbf I}}_l$ is supplied to the CaptionBot, which
produces a captioning sentence $\hat{{\mathbf S}}_l$;
mathematically, we can represent this process by
\begin{eqnarray}
\hat{{\mathbf S}}_l={\cal{C}}(\hat{{\mathbf I}}_l|\theta_{cap})
\end{eqnarray}
where ${\cal{C}}(\cdot)$ denotes the CaptionBot as a nonlinear
operator parameterized by $\theta_{cap}$; conceptually, operator
${\cal{C}}(\cdot)$ can be considered as an approximate inverse of
operator ${\cal{D}}(\cdot)$. Similarly in the right branch, we
have $\hat{{\mathbf S}}_r={\cal{C}}({\mathbf I}^*|\theta_{cap})$
and $\hat{{\mathbf I}}_r={\cal{D}}(\hat{{\mathbf
S}}_r|\theta_{draw})$.

Ideally, we intend to minimize both $\mathbb{E}_{({\mathbf S}^*, {\mathbf I}^*)} \left[L_l(\theta_{draw},\theta_{cap}) \right]$ and $\mathbb{E}_{({\mathbf S}^*, {\mathbf I}^*)} \left[L_r(\theta_{draw},\theta_{cap}) \right]$, in which $L_l(\theta_{draw},\theta_{cap})$  and $L_r(\theta_{draw},\theta_{cap})$ are the loss function of the left hand side and the right hand side of Fig.~\ref{fig:turbo}, respectively. We solve this multi-objective optimization problem by converting it to a single objective optimization problem by summing up these two loss functions, i.e.,
\begin{eqnarray}
\min_{\theta_{draw},\theta_{cap}}  \mathbb{E}_{({\mathbf S}^*, {\mathbf I}^*)} \left[L_l(\theta_{draw},\theta_{cap}) \right] + \mathbb{E}_{({\mathbf S}^*, {\mathbf I}^*)} \left[ L_r(\theta_{draw},\theta_{cap}) \right].
\label{eqn:Lagrangian}
\end{eqnarray}
Our turbo training is intended to solve the optimization problem in \eqref{eqn:Lagrangian} by the stochastic gradient descent algorithm, where we randomly select one of the two terms and approximate the expectation by an empirical average over a mini-batch.  In practice, we select these two terms alternatively, and use the same mini-batch to compute the gradient for both $L_l$ and $L_r$. This reuse of samples cuts the I/O cost by half during the training.



\subsection{Insight of turbo learning}

Later in Section~\ref{subsec:ExperimentalResults}, the
experimental results show that our proposed turbo joint training
of CaptionBot and DrawingBot achieves significant gains over
separate training of CaptionBot and DrawingBot. The reason is as
follows. If we train CaptionBot and DrawingBot jointly under the
turbo approach, the turbo CaptionBot has both sentence information
(generated sentence $\hat{{\mathbf S}}$ vs. gold sentence
${\mathbf S}^*$) and image information (generated image
$\hat{{\mathbf I}}$ vs. gold image ${\mathbf I}^*$) to guide the
update of $\theta_{cap}$; in other words, the turbo CaptionBot is
trained to generate a sentence, which not only is close to the
ground truth sentence, but also captures as many classes/objects
in the original input image as possible. In contrast, the baseline
CaptionBot only has sentence information (generated sentence
$\hat{{\mathbf S}}$ vs. gold sentence ${\mathbf S}^*$) for
updating $\theta_{cap}$. The same reasoning is also true for the
turbo DrawingBot.

\section{Turbo learning for supervised training}
\label{sec:supervised training} This section is organized as
follows. Subsection~\ref{subsec:captionbot} and
Subsection~\ref{subsec:drawingbot} describe CaptionBot and
DrawingBot, respectively; Subsection~\ref{subsec:supervised
learning} describes turbo joint training of CaptionBot and
DrawingBot. Subsection~\ref{subsec:ExperimentalResults} shows the
experimental results.

\subsection{LSTM model for CaptionBot} \label{subsec:captionbot}
\begin{figure}[ht]
    \centering
    \includegraphics[width=0.9\textwidth]{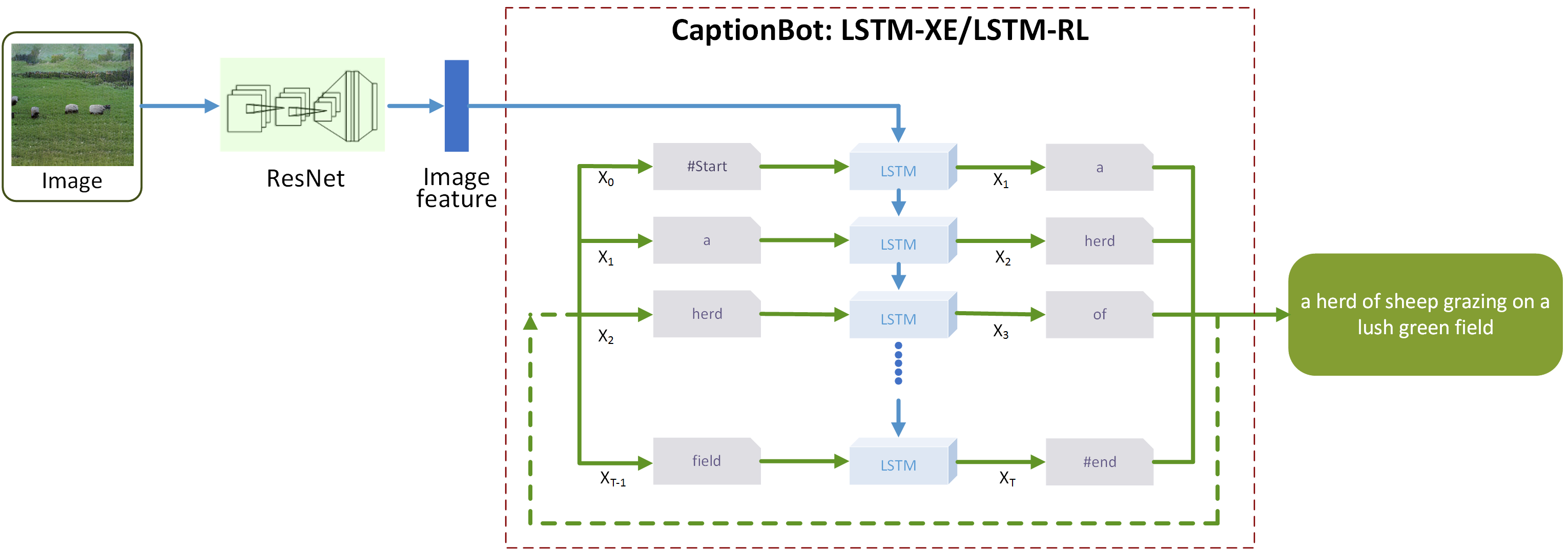}
    \caption{Baseline LSTM model for CaptionBot}
    \label{fig:LSTM}
\end{figure}

To illustrate the benefit of the turbo structure, in this section,
we choose a simple baseline LSTM model as shown in
Fig.~\ref{fig:LSTM} for the CaptionBot .

Consider an image ${\mathbf I}$ with caption ${\mathbf S}$.
Assume that caption ${\mathbf S}$ consists of $T$ words. We define
${\mathbf S}=[{\mathbf x}_1,\cdots , {\mathbf x}_T]$, where
${\mathbf x}_t$ is a one-hot encoding vector of dimension $V$, where
$V$ is the size of the vocabulary. The length $T$ may be different
for different captions. The $t$-th word in a caption, ${\mathbf
x}_t$, is embedded into an $n_x$-dimensional real-valued vector
${\mathbf w}_t={\mathbf W}_e {\mathbf x}_t$, where ${\mathbf W}_e
\in \mathbb{R}^{n_x\times V}$ is a word embedding matrix.

The following is the basic LSTM model for captioning a single
image which is first processed by a Convolutional Neural Network (CNN) such as ResNet and then conveyed to the LSTM \cite{wu2016value,you2016image,SCN_CVPR2017}.
\begin{eqnarray}
{\mathbf h}_t, {\mathbf c}_t = LSTM({\mathbf x}_{t-1},{\mathbf
h}_{t-1}, {\mathbf c}_{t-1}) \label{eqn:basicLSTM}
\end{eqnarray}
where ${\mathbf h}_t, {\mathbf c}_t$ are the hidden state and the
cell state of the LSTM at time $t$, respectively.

The LSTM model in Eq.~\eqref{eqn:basicLSTM} for image captioning
can be trained via minimizing the cross-entropy (XE) loss
function, which is called Baseline LSTM-XE in this paper.


To directly optimize NLP metrics such as CIDEr
\cite{vedantam2015cider} and address the exposure bias issue,
reinforcement learning (RL) can be used to train the LSTM model in
Eq.~\eqref{eqn:basicLSTM} \cite{Rennie2016Self}, which is called
Baseline LSTM-RL in this paper. Under the RL terminology, the LSTM
is regarded as an ``agent'' that interacts with an external
``environment'' (e.g., words and image features). The parameters
of the LSTM, $\theta$, define a policy $p_{\theta}$, that produces
 an ``action'' (e.g., prediction of the current word). After
each action, the agent updates its internal ``state'' (e.g., cell
and hidden states of the LSTM). Upon generating the
end-of-sentence (EOS) symbol, the agent receives a ``reward'' denoted as $r(S)$ (e.g., the CIDEr metric of the generated sentence w.r.t. the
ground-truth sentence), where $S$ is the generated sentence.

Minimizing the reinforcement loss (or maximizing rewards) does not
ensure the readability and fluency of the generated caption
\cite{Pasunuru2017Reinforced}. Using a mixed loss function, which
is a weighted combination of the cross-entropy (XE) loss $L_{XE}$
and the reinforcement learning (RL) loss $L_{RL}$, helps improve
readability and fluency since the cross-entropy loss is based on a
conditioned language model, with a goal of producing fluent
captions. That is, the mixed loss function is given by
\begin{eqnarray}
L_{mix}(\theta) = \gamma L_{RL}(\theta) +(1-\gamma) L_{XE}(\theta)
\label{eqn:mixedLoss}
\end{eqnarray}
where $\gamma\in [0,1]$.

\subsection{AttnGAN for DrawingBot} \label{subsec:drawingbot}

AttnGAN~\cite{Tao18attngan} introduces an attentional multi-stage generative network, which can synthesize fine-grained details at different sub-regions of the image by paying attentions to the relevant words in the natural language description. Trained by multi-level discriminators and a deep attentional multimodal similarity model (DAMSM) that computes a fine-grained image-text matching loss, the AttnGAN achieves the state-of-the-art performance on the text-to-image generation tasks. In this work, we use AttnGAN, more precisely, its attentional multi-stage generative network, as the drawing bot.

\begin{figure}[ht]
    \centering
    \includegraphics[width=0.95\textwidth]{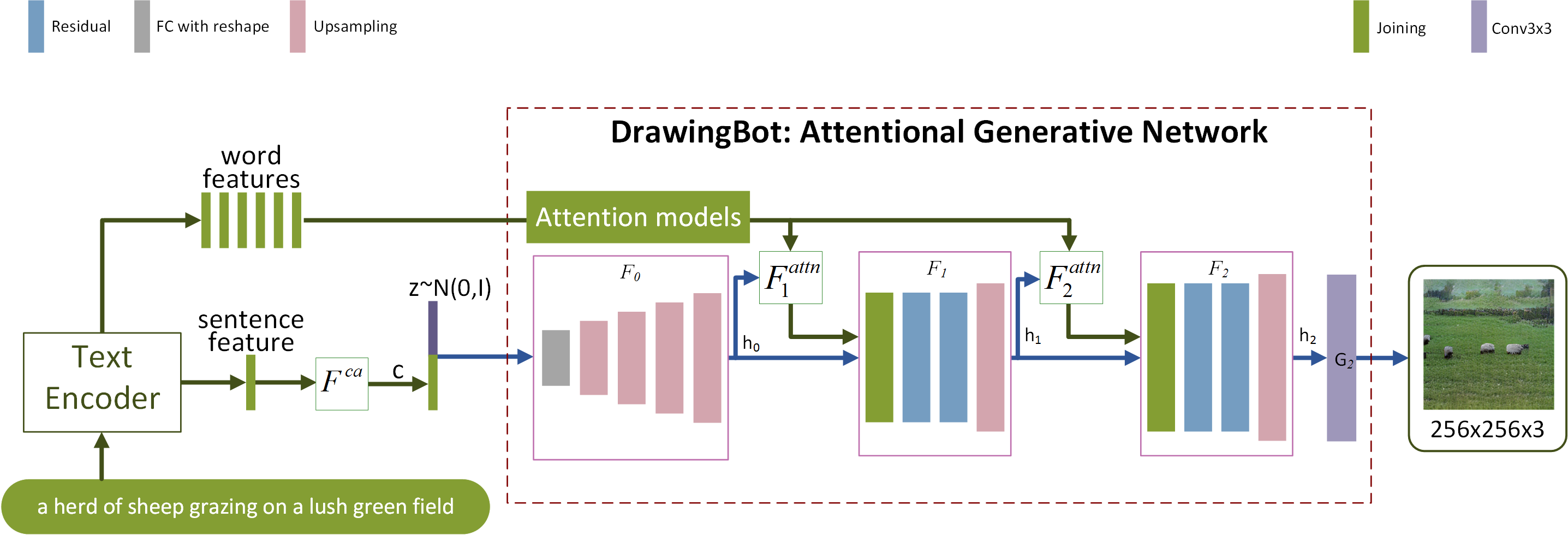}
    \caption{Drawingbot: the attentional multi-stage generative model from AttnGAN~\cite{Tao18attngan} }
    \label{fig:dbot}
\end{figure}

As shown in Figure~\ref{fig:dbot}, the proposed attentional generative network has $m$ stages $F_i$ ($i=0,1,\dots, m$), each of which outputs an hidden state $h_i$. On the last stage, we obtain the generated high-resolution image after passing the last hidden state $h_{m-1}$ through a convolutional layer with $3\times 3$ kernel size. Specifically, the drawing bot, denoted as $\hat{\mathbf I} = \mathcal{C}(\mathbf{S}, z)$, is decomposed as
  \begin{equation}\label{eq:hiddenGs}
   \begin{aligned}
    &h_0 = F_0(z, F^{ca}(\overline{e})); \\ &h_i = F_i(h_{i-1}, F_{i}^{attn}(e, h_{i-1})) \; \text{for} \; i = 1,2, ..., m-1; \; \\ &\hat{\mathbf I} = G_{m-1}(h_{m-1}).
   \end{aligned}
  \end{equation}
Here, $z$ is a noise vector usually sampled from a standard normal distribution.   $\overline{e}$ is the global sentence vector of input sentence $\mathbf{S}$, and $e$ is the matrix of word vectors. The embedding was pretrained by the DAMSM model proposed by \cite{Tao18attngan}. $F^{ca}$ represents the Conditioning Augmentation~\cite{Han16stackgan} that converts the sentence vector $\overline{e}$ to the conditioning vector.   $F^{attn}_i$ is the proposed attention model at the $i^{th}$ stage in \cite{Tao18attngan}. $F^{ca}$, $F^{attn}_i$, $F_i$, and $G_{m-1}$ are all modeled as neural networks.
The AttnGAN is trained by minimizing the sum of 1) the GAN matching loss that jointly approximates
conditional and unconditional distributions of multi-scale images, and 2) a word-level image-text matching loss.

\subsection{Turbo training procedure for supervised learning}
\label{subsec:supervised learning}
Now, we describe the proposed turbo training procedure.
As mentioned before, we select $L_l(\theta_{draw},\theta_{cap})$ and $L_r(\theta_{draw},\theta_{cap})$ alternatively, i.e., minimizing $L_l(\theta_{draw},\theta_{cap})$ and $L_r(\theta_{draw},\theta_{cap})$ alternatively.
Hence, each iteration of the turbo training procedure consists of
the following three steps:
\begin{itemize}
\item {\bf Step 1: minimizing $L_l(\theta_{draw},\theta_{cap})$.}
As shown in the left hand side of Fig.~\ref{fig:turbo}, for a
given gold sentence ${\mathbf S}^*$, the DrawingBot generates an
image $\hat{{\mathbf I}}_l$; the generated image is supplied to
the CaptionBot, which produces a captioning sentence
$\hat{{\mathbf S}}_l$. Then we use the following loss function to
calculate the gradients and update $\theta_{draw}$ and
$\theta_{cap}$ simultaneously:
\begin{eqnarray}
L_l(\theta_{draw},\theta_{cap}) = \beta_2 (\alpha(\hat{{\mathbf I}}_l,
{\mathbf I}^*) - r(\hat{{\mathbf S}}_l)) +(1-\beta_2) (\beta_1
\alpha(\hat{{\mathbf I}}_l, {\mathbf I}^*))+ (1-\beta_1)
L_{XE}(\theta_{cap})),
\label{eqn:Step1_XE_joint}\\
L_l(\theta_{draw},\theta_{cap}) = \beta_2 (\alpha(\hat{{\mathbf I}}_l,
{\mathbf I}^*) - r(\hat{{\mathbf S}}_l)) +(1-\beta_2) (\beta_1
\alpha(\hat{{\mathbf I}}_l, {\mathbf I}^*))+ (1-\beta_1)
L_{mix}(\theta_{cap})), \label{eqn:Step1_RL_joint}
\end{eqnarray}
where $\beta_1,\beta_2\in [0,1]$; \eqref{eqn:Step1_XE_joint} and
\eqref{eqn:Step1_RL_joint} are for the CaptionBot being LSTM-XE
and LSTM-RL, respectively; $r(\hat{{\mathbf S}}_l)$ is a reward
(e.g., the CIDEr metric of the generated sentence $\hat{{\mathbf
S}}_l$ w.r.t. the ground-truth sentence ${\mathbf S}^*$);
$\alpha(\hat{{\mathbf I}}_l, {\mathbf I}^*)$ is $KL(p(y|{\mathbf
I}^*)||p(y|\hat{{\mathbf I}}_l))$ where $p(y|{\mathbf I}^*)$ is
the conditional distribution of label $y$ given the ground-truth
image ${\mathbf I}^*$ and $p(y|\hat{{\mathbf I}}_l)$ is the
conditional distribution of label $y$ given the generated image
$\hat{{\mathbf I}}_l$, and $KL(p||q)$ is the Kullback-Leibler
divergence from $q$ to $p$;  label $y$ is generated by
Inception-v3 model in TensorFlow \cite{tensorflow2015-whitepaper}.
A label $y$ is selected from 1,000 meaningful classes, such as
``Zebra'' and ``Dishwasher''.

\item {\bf Step 2: minimizing $L_r(\theta_{draw},\theta_{cap})$.}
As shown in the right hand side of Fig.~\ref{fig:turbo}, for a
given gold image ${\mathbf I}^*$, the CaptionBot generates a
sentence $\hat{{\mathbf S}}_r$; the generated sentence is supplied
to the DrawingBot, which produces an image $\hat{{\mathbf I}}_r$.
Then we use the loss function $L_r(\theta_{draw},\theta_{cap})$,
which is defined by replacing $(\hat{\mathbf S}_l, \hat{\mathbf
I}_l)$ with $(\hat{\mathbf S}_r, \hat{\mathbf I}_r)$ in
\eqref{eqn:Step1_XE_joint} or \eqref{eqn:Step1_RL_joint}, to
calculate the gradients and update $\theta_{draw}$ and
$\theta_{cap}$ simultaneously.

\item {\bf Step 3:} Go to Step 1 until convergence.
\end{itemize}

{  Note that it is critical to choose the right reconstruction
loss to enforce cycle consistency in turbo learning. After
experimenting with pixel-wise loss,  Deep  Structured  Semantic
Models  (DSSM) loss \cite{Huang:2013:LDS:2505515.2505665}, and our
``perceptual''-like loss $\alpha(\hat{{\mathbf I}}_l, {\mathbf
I}^*)$, we found that our ``perceptual''-like loss achieves
significantly better performance.

It is worth mentioning that the above turbo training is different
from GAN training. GAN training is adversarial while our turbo
training is collaborative. Actually, our method is more similar to
the training of Auto-Encoder (AE). Each branch of the turbo
training in Fig.~\ref{fig:turbo}) can be viewed as an AE. For
example, the left branch is an AE where the DrawingBot is the
encoder and the CaptionBot is the decoder. Different from vinilla
AE (VAE), our encoding space is semantically meaningful, and
supervised signal is available for the encoding procedure because
image-caption pairs are available. For the decoder/reconstruction
loss, instead of using pixel-wise loss in VAE, we propose to use a
``perceptual''-like loss to better capture its semantics.}

\subsection{Experimental results}
\label{subsec:ExperimentalResults}

\subsubsection{Dataset}\label{subsubsec:Dataset}

To evaluate the performance of our proposed approach, we use the
COCO dataset \cite{COCO_weblink}. The COCO dataset contains
123,287 images, each of which is annotated with at least 5
captions. We use the same pre-defined splits as in
\cite{karpathy2015deep,SCN_CVPR2017}: 113,287 images for training,
5,000 images for validation, and 5,000 images for testing. We use
the same vocabulary as that employed in \cite{SCN_CVPR2017}, which
consists of 8,791 words.

\begin{table}[th]
  \caption{Performance of CaptionBot and corresponding DrawingBot with BLEU-4 as the reward $r(\hat{{\mathbf S}})$.}
  \label{table:BLEU_100000}
  \centering
{\small  \begin{tabular}{llllll||ll}
    \toprule
Captionbot     &  BLEU-4 & CIDEr-D & ROUGE-L & METEOR &  SPICE & Drawingbot & Inception\\
    \midrule
{\scriptsize Baseline LSTM-XE} &   0.2684 &  0.7362 & 0.4937 & 0.2193 & 0.1663 &{\scriptsize  AttnGAN} & 25.68\\
{\scriptsize Turbo LSTM-XE} &\textbf{0.3168}   & \textbf{0.7364}&   \textbf{0.4938}& \textbf{0.2196}& \textbf{0.1720}& {\scriptsize Turbo AttnGAN} & \textbf{26.69}\\
\midrule
{\scriptsize Baseline LSTM-RL} &  0.2831  & 0.7238 & 0.4963 & 0.2190 & 0.1733 &{\scriptsize AttnGAN} & 25.68\\
{\scriptsize Turbo LSTM-RL}  & \textbf{0.3183} &\textbf{0.7327} &\textbf{0.4974}&    \textbf{0.2191}& \textbf{0.1735} &{\scriptsize Turbo AttnGAN}  & \textbf{26.88}\\
    \bottomrule
  \end{tabular}
}
\end{table}

\subsubsection{Evaluation}\label{subsec:Evaluation}

For the CNN, which is used to extract features from an image, we
used ResNet-50 \cite{he2016deep} pretrained on the ImageNet
dataset. The reason to use ResNet-50 instead of ResNet-152 is mainly for the consideration of training and inference efficiency.  We will report the experimental results for ResNet-152 in our future work.

The feature vector ${\mathbf v}$ has 2048 dimensions.
Word embedding vectors in ${\mathbf W}_e$ are downloaded from the
web \cite{Stanford_Glove_weblink}. The model is implemented in
TensorFlow \cite{tensorflow2015-whitepaper} with the default
settings for random initialization and optimization by
backpropagation.  We empirically set $\beta_1=\beta_2=0.5$.

The widely-used BLEU \cite{papineni2002bleu}, METEOR
\cite{banerjee2005meteor}, CIDEr \cite{vedantam2015cider}, and
SPICE \cite{anderson2016spice} metrics are reported in our
quantitative evaluation of the performance of the proposed
approach.

Table~\ref{table:BLEU_100000} shows the experimental results on
the COCO dataset with 113,287 training samples, for which we use
BLEU-4 as the reward $r(\hat{{\mathbf S}})$ in
\eqref{eqn:Step1_XE_joint}, and \eqref{eqn:Step1_RL_joint}. As
shown in the table, our proposed turbo approach achieves
significant gain over separate training of CaptionBot and
DrawingBot. Following~\cite{Tao18attngan}, we use the  Inception
score \cite{Salimans2016Improved} as the quantitative evaluation
measure.  The larger Inception score, the better performance.
Table~\ref{table:BLEU_100000} shows the Inception score for
AttnGAN and Turbo AttnGAN.  It is observed that turbo AttnGANs
achieve higher Inception scores than AttnGAN.

Table~\ref{table:CIDEr_100000} shows the experimental results on
the COCO dataset with 113,287 training samples, for which we use
CIDEr-D as the reward $r(\hat{{\mathbf S}})$ in
\eqref{eqn:Step1_XE_joint}, and \eqref{eqn:Step1_RL_joint}. As
shown in the table, our proposed turbo approach achieves
significant gain over separate training of CaptionBot and
DrawingBot.

\begin{table}[th]
\caption{Performance of CaptionBot and corresponding DrawingBot
with CIDEr-D as the reward $r(\hat{{\mathbf S}})$.}
  \label{table:CIDEr_100000}
  \centering
{\small  \begin{tabular}{llllll||ll}
    \toprule
Captionbot     &  BLEU-4 & CIDEr-D & ROUGE-L & METEOR &  SPICE & Drawingbot & Inception\\
    \midrule
{\scriptsize Baseline LSTM-XE} &   0.2684 &  0.7362 & 0.4937 & 0.2193 & 0.1663 & {\scriptsize AttnGAN} & 25.68\\
{\scriptsize Turbo LSTM-XE} &\textbf{0.3033}&    \textbf{0.7473}& \textbf{0.4942}& \textbf{0.2195}& \textbf{0.1721}& {\scriptsize Turbo AttnGAN} & \textbf{26.72}\\
\midrule
{\scriptsize Baseline LSTM-RL} &  0.2831  & 0.7238 & 0.4963 & 0.2190 & 0.1733 &{\scriptsize AttnGAN} & 25.68\\
{\scriptsize Turbo LSTM-RL}  & \textbf{0.3162}&  \textbf{0.7478} &    \textbf{0.4981}& \textbf{0.2192}& \textbf{0.1748} &{\scriptsize Turbo AttnGAN}  & \textbf{26.83}\\
    \bottomrule
  \end{tabular}
}
\end{table}

Because we use ResNet-50 instead of ResNet-152, the performance of
the baseline LSTM CaptionBot is not as good as the
state-of-the-art LSTM, which uses ResNet-152 or better features.

It is worth mentioning that this paper aims at developing a turbo
learning approach to training LSTM-based CaptionBot; therefore, it
is directly comparable to an LSTM baseline. Therefore, in the
experiments, we focus on the comparison to a strong CNN-LSTM
baseline. We acknowledge that more recent papers
\cite{xu2017image,Rennie2016Self,yao2016boosting,lu2017knowing,SCN_CVPR2017}
reported better performance on the task of image captioning.
Performance improvements in these more recent models are mainly
due to using better image features such as those obtained by
region-based convolutional neural networks (R-CNN), or using more
complex attention mechanisms \cite{SCN_CVPR2017} to provide a
better context vector for caption generation, or using an ensemble
of multiple LSTMs, among others. However, the LSTM is still
playing a core role in these works and we believe improvement over
the core LSTM by turbo learning is still very valuable and
orthogonal to most existing works; that is why we compare the
turbo LSTM with a native LSTM.

\section{Semi-supervised learning for CaptionBot}
\label{sec:Semi-supervised}
\subsection{Semi-supervised learning approach for CaptionBot}
In this section, we present a semi-supervised learning approach
for training CaptionBot.

Human annotation is costly.  Hence, many images on the Internet
has no caption and semi-supervised learning for CaptionBot is
desirable. In fact, the right hand side of Fig.~\ref{fig:turbo}
forms a loop, which enables semi-supervised learning for
CaptionBot since the closed loop can provide pseudo-labels for
unlabeled samples.

As shown in the right hand side of Fig.~\ref{fig:turbo}, for a
given gold image ${\mathbf I}^*$, the CaptionBot generates a
sentence $\hat{{\mathbf S}}$; the generated sentence is supplied
to the DrawingBot, which produces an image $\hat{{\mathbf I}}$.
The generated image $\hat{{\mathbf I}}$ is supplied to the
CaptionBot, which generates a sentence $\tilde{{\mathbf S}}$.

Then we can use the following equation to calculate the loss for an
unlabeled sample:
\begin{eqnarray}
L(\theta_{cap}) = \beta_1 \times \alpha(\hat{{\mathbf I}},
{\mathbf I}^*))+ (1-\beta_1) \times  L_{XE}(\theta_{cap}) \qquad
\mbox{for LSTM-XE CaptionBot,} \label{eqn:Step2-1_semi}\\
L(\theta_{cap}) = \beta_1 \times \alpha(\hat{{\mathbf I}},
{\mathbf I}^*))+ (1-\beta_1) \times  CIDEr(\hat{{\mathbf
S}},\tilde{{\mathbf S}}) \qquad \mbox{for LSTM-RL CaptionBot,}
\label{eqn:Step2-2_semi}
\end{eqnarray}
where $\beta_1\in [0,1]$; $\alpha(\hat{{\mathbf I}}, {\mathbf
I}^*)$ is $KL(p(y|{\mathbf I}^*)||p(y|\hat{{\mathbf I}}))$, and
$L_{XE}(\theta_{cap})$ is the cross-entropy between $\hat{{\mathbf
S}}$ and $\tilde{{\mathbf S}}$; $CIDEr(\tilde{{\mathbf
S}},\hat{{\mathbf S}})$ is the CIDEr metric of the sentence
$\tilde{{\mathbf S}}$ w.r.t. the sentence $\hat{{\mathbf S}}$.

In our experiments, in order to smoothly transit from the initial
model trained from labeled data to the model training from labeled
and unlabeled data, we adopted the following strategy. For each
mini-batch, we use a half number of samples from unlabeled data
and another half number of samples from labeled data (sampled from
the labeled dataset used to train the initial model). The
objective is to minimize the total loss of all labeled and
unlabeled samples. We compute the gradient of the total loss  of
all labeled and unlabeled samples in a mini-batch and update the
neural weights $\theta_{cap}$ of the CaptionBot.

{  Here, we would like to relate our semi-supervised learning to
the CycleGAN scheme \cite{CycleGAN2017}. Both schemes use cycle
consistency, but they are implemented in different ways.
Specifically, CycleGAN uses pixel-wise $L_1$ loss while we use
``perceptual''-like loss for image semantic consistency. CycleGAN
has not been applied to text reconstruction, while we use CIDEr
score for caption consistency. While CycleGAN was mainly used in
homogeneous modality for image-to-image translation, our framework
works for multi-modality problems, i.e., image-to-text and
text-to-image tasks. }

\subsection{Performance Evaluation}

Same as Section~\ref{subsec:ExperimentalResults}, we also use COCO
dataset to evaluate the performance of the proposed
semi-supervised learning approach for CaptionBot.  We set
$\beta_1=0.85$.

Table~\ref{table:CIDEr_1000label} shows the experimental results
for semi-supervised learning with 1,000 labeled training samples,
112,287 unlabeled training samples and CIDEr-D as the reward.  It
is observed that turbo LSTMs significantly outperform baseline LSTMs for semi-supervised image captioning.

\begin{table}[th]
\caption{Performance of CaptionBot with 1,000 labeled training
samples, 112,287 unlabeled training samples and CIDEr-D as the
reward $r(\hat{{\mathbf S}})$.}
  \label{table:CIDEr_1000label}
  \centering
{\small  \begin{tabular}{llllll}
    \toprule
Methods     &  BLEU-4 & CIDEr-D & ROUGE-L & METEOR &  SPICE\\
    \midrule
Baseline LSTM-XE &   0.1036 & 0.4331 & 0.3462 &  0.1864 & 0.1231  \\
Turbo LSTM-XE & \textbf{0.1832} &    \textbf{0.6841} &    \textbf{0.4174} &    \textbf{0.2071}& \textbf{0.1432}\\
\midrule
Baseline LSTM-RL &  0.1181   & 0.4924 & 0.3518 & 0.1863 & 0.1361 \\
Turbo LSTM-RL  & \textbf{0.1982} & \textbf{0.6954} &  \textbf{0.4215} &    \textbf{0.2082} &    \textbf{0.1466} \\
    \bottomrule
  \end{tabular}
}
\end{table}

\begin{figure}[ht]
    \centering
    \includegraphics[width=\textwidth]{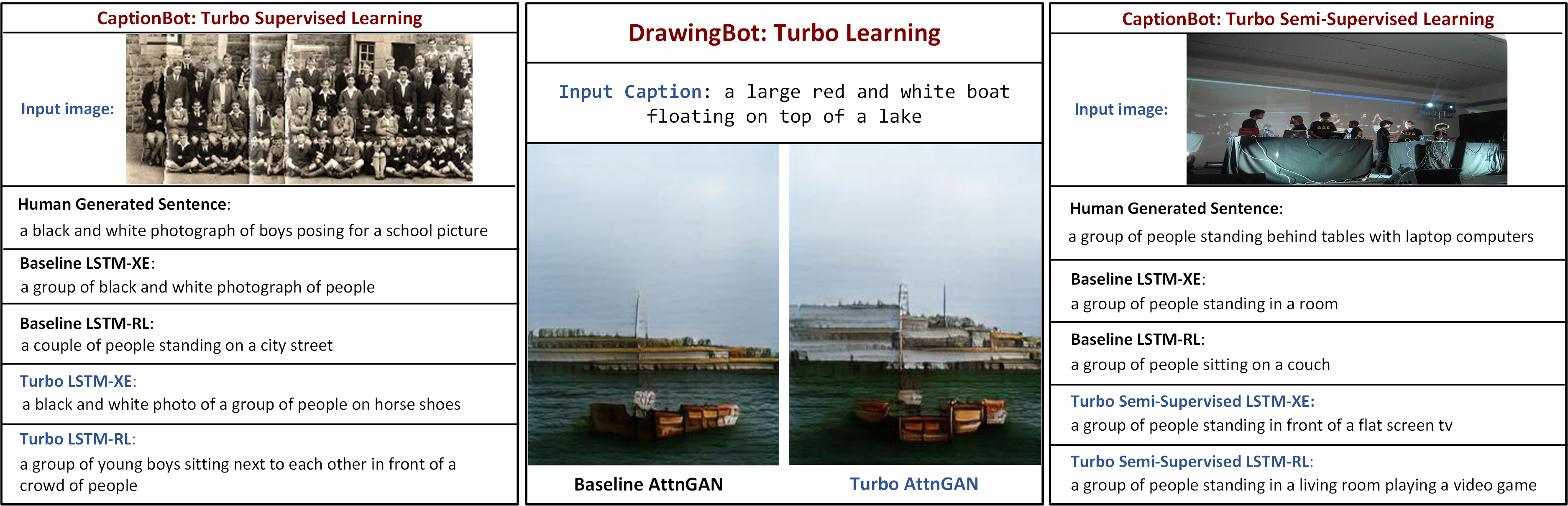}
    \caption{Sample results of CaptionBot and DrawingBot for supervised and semi-supervised learning}
    \label{fig:drawingbot1}
\end{figure}

Fig.~\ref{fig:drawingbot1} shows some sample results of CaptionBot and DrawingBot for supervised and semi-supervised learning.

\section{Conclusion}
\label{sec:Conclusion}

We have presented a novel turbo learning approach to jointly
training a CaptionBot and a DrawingBot. To the best of our
knowledge, this is the first work that studies both problems in
one framework. The framework leverages the duality of image
captioning and text-to-image generation and forms a closed loop,
which results in a new loss metric by comparing the initial input
with the feedback produced by the whole loop. This not only leads
to better CaptionBot and DrawingBot by joint training, but also
makes semi-supervised learning possible. Experimental results on
the COCO dataset have effectively validated the advantages of the
proposed joint learning approach.

In our future work, we will explore if adding more unlabeled data can further improve the performance of both bots, and extend the turbo learning approach to other domains, for example,
speech recognition vs. text to speech, question answering vs.
question generation, search (matching queries to
documents) vs. keyword extraction (extracting keywords/queries
for documents).


\small
\bibliographystyle{IEEEtran}
\bibliography{journal,paperbib}

\clearpage
\section*{Supplementary Materials}

\begin{figure}[h]
      \centering
      \includegraphics[width=\textwidth]{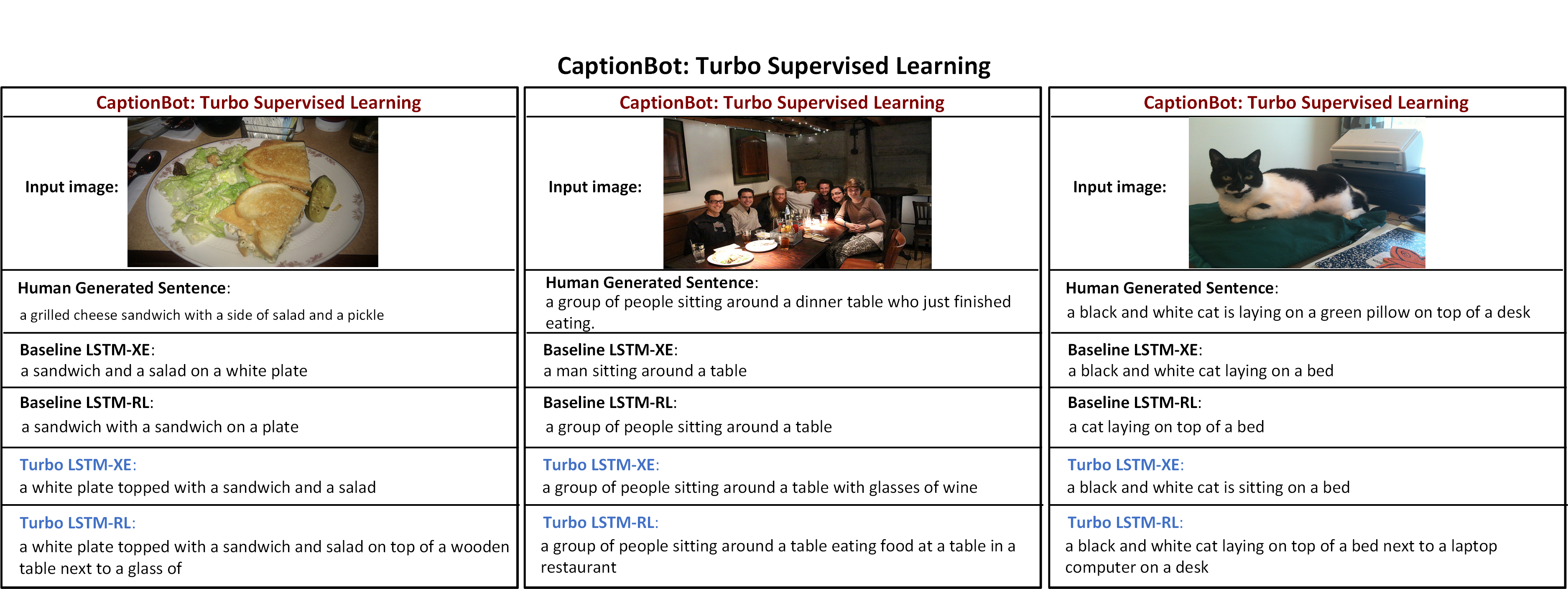}
      \caption{Sample results of CaptionBot for supervised learning with CIDEr-D as the reward $r(\hat{{\mathbf S}})$}
      \label{fig:captionbot}
  \end{figure}

\begin{figure}[h]
      \centering
      \includegraphics[width=\textwidth]{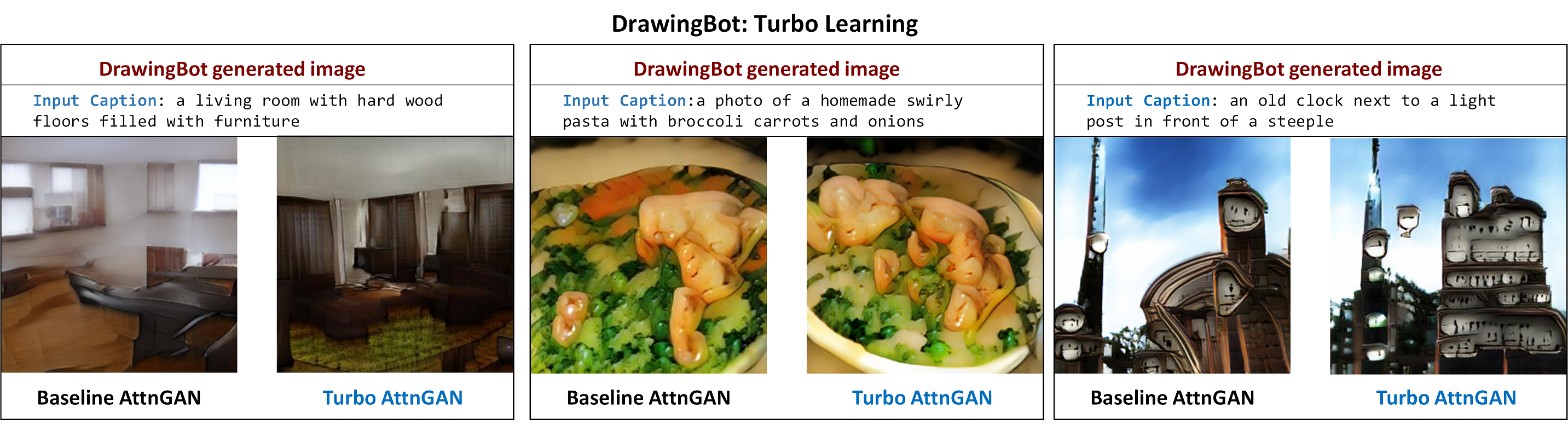}
      \caption{Sample results of DrawingBot for supervised learning with CIDEr-D as the reward $r(\hat{{\mathbf S}})$}
      \label{fig:drawingbot2}
  \end{figure}

\begin{figure}[h]
      \centering
      \includegraphics[width=\textwidth]{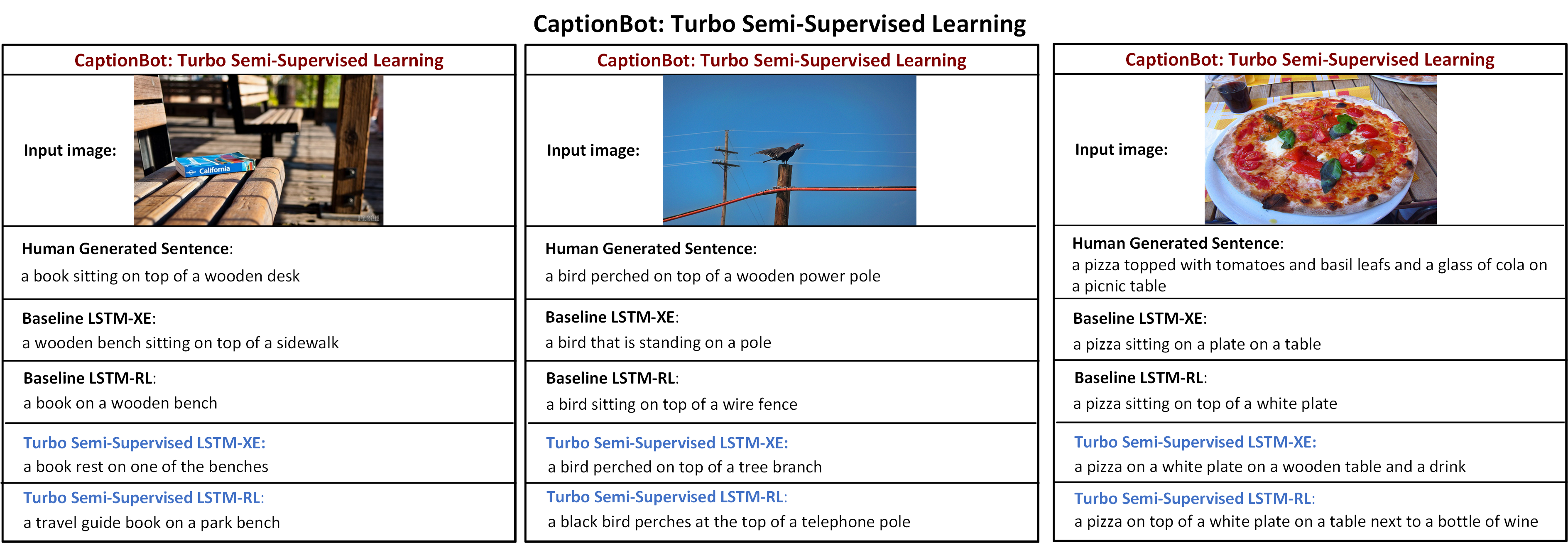}
      \caption{Sample results of CaptionBot for semi-supervised learning with CIDEr-D as the reward $r(\hat{{\mathbf S}})$}
      \label{fig:captionbot-semi}
  \end{figure}

\begin{itemize}[leftmargin=20pt]
\item  Fig.~\ref{fig:captionbot} shows some sample results of CaptionBot in the supervised setting.
\item Fig.~\ref{fig:drawingbot2} shows some sample results of DrawingBot in the supervised setting.
\item Fig.~\ref{fig:captionbot-semi} shows some sample results of CaptionBot in the semi-supervised setting.
\end{itemize}

\end{document}